\documentclass[conference]{IEEEtran}
\IEEEoverridecommandlockouts
\usepackage{amsmath,amsfonts}
\usepackage{algorithmic}
\usepackage{array}
\usepackage[ruled,vlined]{algorithm2e}
\usepackage{breqn}
\usepackage[caption=false,font=normalsize,labelfont=sf,textfont=sf]{subfig}
\usepackage{textcomp}
\usepackage{cite}
\usepackage{hyperref}
\hypersetup{
    colorlinks = true,
    allcolors = blue
}
\usepackage{multirow}
\usepackage{stfloats}
\renewcommand{\arraystretch}{1.5} 
\usepackage{url}
\usepackage{pgfplots}
\usepackage{verbatim}
\usepackage{pgfplotstable}
\usepackage{graphicx}
\usepackage{tikz}
\usepackage{color,soul} 
\usepackage{xcolor}
\def\BibTeX{{\rm B\kern-.05em{\sc i\kern-.025em b}\kern-.08em
    T\kern-.1667em\lower.7ex\hbox{E}\kern-.125emX}}
\begin{document}

\title{\normalsize{\textcolor{blue}{\bf This paper was accepted to IEEE MWSCAS 2026}}\\ \large \bf QFedAgent: Quantum-Enhanced Personalized Federated Learning for Multi-Agent Activity Recognition}\vspace{-0.5cm}
%\title{\LARGE{\bf QFedAgent: Quantum-Enhanced Personalized Federated Learning for Multi-Agent Activity Recognition}}\vspace{-0.2cm}

\author{\IEEEauthorblockN{Quoc Bao Phan and Tuy Tan Nguyen}
\IEEEauthorblockA{\textit{Department of Electrical and Computer Engineering} \\\textit{FAMU-FSU College of Engineering, Florida State University}\\
Tallahassee, FL 32310, USA \\
\{qp25c, tuy.nguyen\}@fsu.edu}
}

\maketitle

\begin{abstract}
Federated learning (FL) enables collaborative model training across distributed devices without sharing raw data, making it suitable for privacy-sensitive robotic sensing applications. However, multi-agent systems generate heterogeneous and non-independent and identically distributed (non-IID) multimodal sensor streams that degrade conventional FL algorithms, while classical fusion modules introduce substantial parameter overhead and communication cost. This paper proposes QFedAgent, a hybrid quantum-classical personalized FL framework for multi-agent activity recognition. The approach integrates a variational quantum circuit fusion module that models accelerometer--gyroscope interactions through quantum state encoding and entanglement, requiring only 72 quantum rotation parameters versus 33K in classical multi-layer perceptron-based fusion, achieving approximately $10\times$ total parameter reduction. Experiments on the OPPORTUNITY dataset under subject-based non-IID partitions demonstrate 97.7\% mean test accuracy, confirming that parameter-efficient quantum fusion remains competitive with conventional federated baselines.
\end{abstract}


\section{Introduction}
Autonomous robotic systems are increasingly deployed in industrial automation, healthcare monitoring, and intelligent infrastructure. In many applications, agents operate as distributed teams equipped with heterogeneous sensing platforms such as inertial measurement units (IMUs), cameras, and LiDAR. As they continuously interact with their environments, these agents generate large volumes of multimodal sensory data that are geographically distributed across devices. However, centralizing such data for model training is often impractical due to communication overhead, latency constraints, and privacy concerns, particularly in safety-critical environments where reliable connectivity cannot be guaranteed \cite{mughal2024adaptive_fl_edge}.

To address these challenges, federated learning (FL) has emerged as a promising framework for collaborative training in distributed robotic systems \cite{10753028}. In FL, agents train models locally on their own observations and share model updates with a coordinating server instead of raw data, preserving data locality while enabling collective learning \cite{electronics14071323}. However, classical methods such as FedAvg generally assume similar client data distributions, whereas robotic agents often operate under different environments, motion patterns, and sensor characteristics, leading to non-independent and identically distributed (non-IID) data that degrades convergence and weakens personalization \cite{10825769}. This issue becomes even more pronounced in multimodal perception tasks such as activity recognition, where IMU-based accelerometer and gyroscope signals exhibit complex temporal and statistical dependencies. Although fusion methods such as attention, gating, and transformers can model these interactions effectively, they usually introduce substantial parameter overhead, increasing communication cost and further complicating aggregation across heterogeneous clients \cite{KHAN2025111844}.

To mitigate these limitations, compact yet expressive fusion mechanisms are desirable for multimodal FL systems. Variational quantum circuits (VQCs) have recently emerged as differentiable models that can be optimized using gradient-based methods \cite{s25041277}. By encoding classical inputs into quantum states and exploiting entanglement operations, VQCs can represent complex feature correlations while maintaining relatively few trainable parameters. Consequently, they offer a promising direction for parameter-efficient multimodal fusion in distributed learning scenarios. However, despite these advantages, integrating quantum circuit layers into personalized FL systems for robotic sensing remains largely unexplored.

We propose QFedAgent, a hybrid quantum–classical personalized FL framework for multi-agent activity recognition from multimodal IMU data. The framework uses VQC as parameter-efficient cross-modal fusion operators while maintaining robust performance under non-IID federated training. Our main contributions are summarized as follows:

\begin{itemize}
    \item We propose a VQC-based fusion module that models accelerometer--gyroscope interactions via quantum state encoding and entanglement, with 72 quantum rotation parameters achieving approximately $10\times$ total parameter reduction over classical MLP-based fusion.
    \item We design a personalized FL architecture with a shared encoder and VQC fusion module aggregated globally while maintaining client-specific adapter and classification heads.
    \item We evaluate the framework on the OPPORTUNITY activity recognition dataset~\cite{roggen2010collecting} under subject-based non-IID partitions, achieving 97.7\% mean test accuracy with substantially lower fusion complexity.
\end{itemize}
\section{System Model and Method}
\label{sec:method}

\subsection{Problem Formulation}
\begin{figure*}[t]
    \centering
    \includegraphics[width=0.65\linewidth]{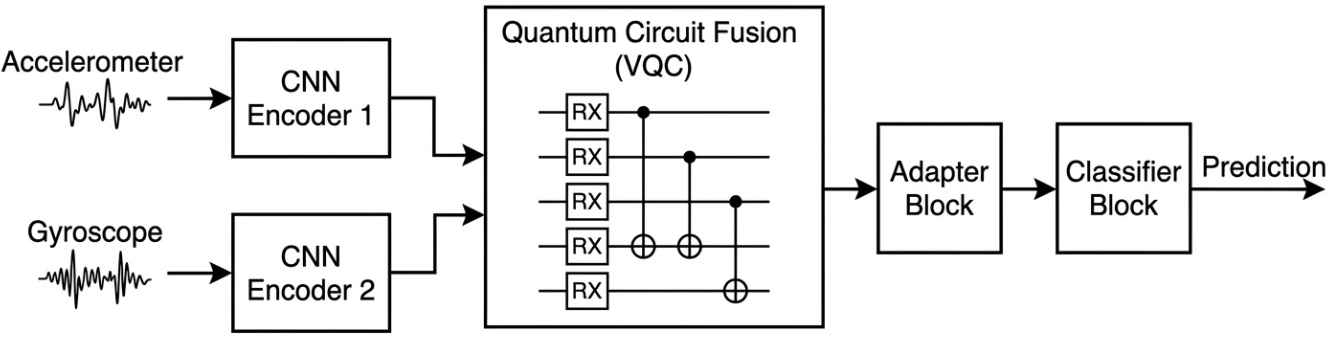}
\caption{QFedAgent architecture: dual CNN encoders feed a VQC fusion layer; adapter and classifier blocks remain local while encoders and VQC weights are aggregated globally.}
\label{fig:architecture}
\end{figure*}

The overall QFedAgent architecture is illustrated in Fig.~\ref{fig:architecture}. Each agent processes accelerometer and gyroscope signals using separate CNN encoders to produce modality embeddings. These embeddings are fused by a shared VQC layer that captures cross-modal interactions through quantum entanglement. The resulting representation is then refined by a client-specific adapter and fed to a local classifier. During federated training, only the encoder and VQC parameters are transmitted to the server for aggregation, while the adapter and classifier remain private to each client.

Formally, consider a federated system with $K$ agents $\{1,\ldots,K\}$, where client $k$ holds a local dataset $\mathcal{D}_k$ of size $n_k$, and $n=\sum_{k=1}^{K}n_k$. Each client observes two IMU modalities: accelerometer streams $\mathbf{X}^a\in\mathbb{R}^{C\times T}$ and gyroscope streams $\mathbf{X}^g\in\mathbb{R}^{C\times T}$, where $C$ denotes the number of sensor channels and $T$ the number of timesteps per observation window. Because agents operate under different sensing conditions and motion patterns, client data distributions $\mathcal{P}_k$ are heterogeneous ($\mathcal{P}_i\neq\mathcal{P}_j$ for $i\neq j$). The global objective is therefore
\begin{equation}
\min_{\mathbf{w}}F(\mathbf{w})=\sum_{k=1}^{K}\frac{n_k}{n}F_k(\mathbf{w}),
\end{equation}
where
\[
F_k(\mathbf{w})=\mathbb{E}_{(\mathbf{x},y)\sim\mathcal{D}_k}\big[\ell(f(\mathbf{x};\mathbf{w}),y)\big]
\]
denotes the local empirical loss under a cross-entropy criterion.

\subsection{Model Architecture}
The pipeline consists of modality encoding, VQC-based multimodal fusion, client-specific adaptation, and federated aggregation of shared parameters. Model parameters are partitioned into global parameters $\mathbf{w}^g$ that are aggregated across clients and local parameters $\mathbf{w}^l_k$ that remain private:
\begin{equation}
\mathbf{w}^g=\{\mathbf{w}^{enc},\boldsymbol{\theta}^{vqc},\mathbf{w}^{proj}\},\qquad
\mathbf{w}^l_k=\{\mathbf{w}^{cls}_k,\mathbf{w}^{adp}_k\}
\end{equation}
where $\mathbf{w}^{enc}$ are shared dual CNN encoder weights, $\boldsymbol{\theta}^{vqc}\in\mathbb{R}^{L\times N\times3}$ are the VQC rotation parameters for $L$ entangling layers and $N$ qubits, $\mathbf{w}^{proj}$ are classical projection weights, $\mathbf{w}^{cls}_k$ is the client-specific classifier head, and $\mathbf{w}^{adp}_k$ is a lightweight adapter that refines fused features to capture client-specific sensing characteristics.

Each modality stream is processed by a shared CNN composed of three convolutional blocks with batch normalization and ReLU activation, followed by adaptive average pooling and a linear projection to a $d$-dimensional embedding space, producing modality embeddings $\mathbf{e}^a,\mathbf{e}^g\in\mathbb{R}^d$.

\subsection{VQC Fusion Layer}
\begin{figure}[t]
\centering
\includegraphics[width=0.975\columnwidth]{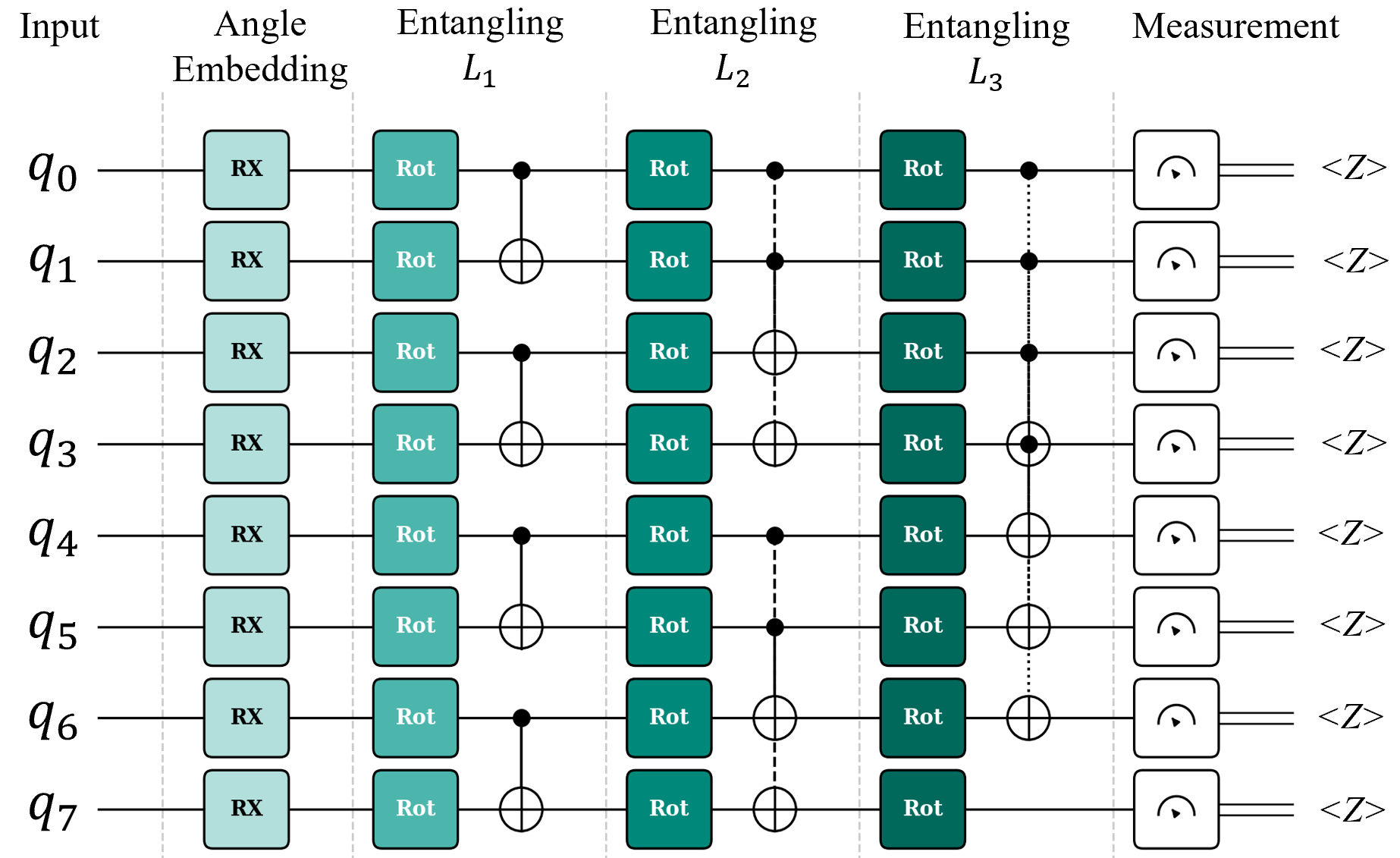}
\caption{VQC fusion layer ($N=8$ qubits, $L=3$). Accelerometer and gyroscope embeddings are encoded on $q_0$--$q_3$ and $q_4$--$q_7$ via RX rotations. Entangling layers create cross-modal correlations, and Pauli-$Z$ measurements produce the fused representation $\mathbf{z}\in\mathbb{R}^N$.}
\label{fig:vqc}
\end{figure}
The two modality embeddings are concatenated and projected into qubit rotation angles:
\begin{equation}
\boldsymbol{\phi}=\pi\cdot\tanh\!\left(\mathbf{W}_{pre}[\mathbf{e}^a\|\mathbf{e}^g]\right)\in[-\pi,\pi]^N
\end{equation}
where $\mathbf{W}_{pre}$ maps the $2d$-dimensional embedding space to $N$ qubits. The angles are encoded via \texttt{AngleEmbedding}, and cross-modal interaction is modeled through $L$ layers of \texttt{StronglyEntanglingLayers} parameterized by $\boldsymbol{\theta}^{vqc}$:
\begin{equation}
U(\boldsymbol{\phi},\boldsymbol{\theta})=\text{SEL}(\boldsymbol{\theta})\cdot\text{AngleEmbed}(\boldsymbol{\phi})
\end{equation}
The fused representation is obtained by measuring Pauli-$Z$ expectations
\begin{equation}
z_i=\langle\psi(\boldsymbol{\phi},\boldsymbol{\theta})|Z_i|\psi(\boldsymbol{\phi},\boldsymbol{\theta})\rangle,\quad i=1,\ldots,N
\end{equation}
producing $\mathbf{z}\in[-1,1]^N$. The final fused embedding is
\begin{equation}
\mathbf{f}=\text{LN}(\mathbf{W}_{post}\mathbf{z})\in\mathbb{R}^d
\end{equation}
where $\mathbf{W}_{post}$ maps measurements back to the embedding space and $\text{LN}(\cdot)$ denotes layer normalization. Since Pauli measurements satisfy $|z_i|\le1$, the VQC fusion layer remains bounded and Lipschitz-continuous, ensuring stable gradient-based optimization. The circuit parameters $\boldsymbol{\theta}^{vqc}$ are optimized via the parameter-shift rule \cite{PhysRevA032331}, enabling unbiased gradient estimates compatible with classical backpropagation. The resulting circuit structure is illustrated in Fig.~\ref{fig:vqc}. In the diagram, RX boxes denote single-qubit rotation gates encoding the input angles, $\bullet$ indicates control qubits, and $\oplus$ indicates the corresponding CNOT targets. Together, these entangling operations create cross-modal correlations between the accelerometer qubits ($q_0$--$q_3$) and gyroscope qubits ($q_4$--$q_7$).

\subsection{Federated Training Protocol}
\begin{algorithm}[!t]
\DontPrintSemicolon
\SetNoFillComment
\caption{QFedAgent: Personalized Federated Learning with Quantum Fusion}
\label{alg:vqc-fl}
\KwIn{Clients $\{1,\ldots,K\}$, datasets $\{\mathcal{D}_k\}$, rounds $R$, learning rate $\eta$}
\KwOut{Global parameters $\mathbf{w}^g_R$, local parameters $\{\mathbf{w}^l_{k,R}\}$}
Initialize $\mathbf{w}^g_0 = \{\mathbf{w}^{enc}, \boldsymbol{\theta}^{vqc}, \mathbf{w}^{proj}\}$ on server;
each client $k$ initializes $\mathbf{w}^l_{k,0} = \{\mathbf{w}^{cls}_k, \mathbf{w}^{adp}_k\}$ \;
\For{each round $r = 1, \ldots, R$}{
  Server broadcasts $\mathbf{w}^g_{r-1}$ to all clients \;
  \For{each client $k$ \emph{in parallel}}{
    \tcp{Local training for one epoch}
    \For{each mini-batch $(\mathbf{X}^a, \mathbf{X}^g, y) \sim \mathcal{D}_k$}{
      Encode $\mathbf{e}^a, \mathbf{e}^g$ via CNN; fuse via VQC to get $\mathbf{f}$ \;
      Classify: $\hat{y} = \mathbf{w}^{cls}_k\!\left(\mathbf{f} + \mathbf{w}^{adp}_k(\mathbf{f})\right)$ \;
      Update $\{\mathbf{w}^g, \mathbf{w}^l_k\} \leftarrow \{\mathbf{w}^g, \mathbf{w}^l_k\} - \eta\,\nabla\ell(\hat{y}, y)$ \;
    }
    Send $\mathbf{w}^g_{k,r}$ and $n_k$ to server 
  }
  \tcp{Weighted FedAvg aggregation}
  $\mathbf{w}^g_r \leftarrow \sum_{k=1}^{K} \dfrac{n_k}{n}\,\mathbf{w}^g_{k,r}$ \;
}
\Return $\mathbf{w}^g_R,\; \{\mathbf{w}^l_{k,R}\}_{k=1}^{K}$
\end{algorithm}
Training proceeds over $R$ communication rounds as detailed in Algorithm~\ref{alg:vqc-fl}. At each round $r$, the server broadcasts the current global parameters $\mathbf{w}^g_r=\{\mathbf{w}^{enc},\boldsymbol{\theta}^{vqc},\mathbf{w}^{proj}\}$ to all clients. Each client performs one local training epoch, updating both global and local parameters, then returns only the updated global parameters to the server. The server aggregates them via weighted FedAvg:
\begin{equation}
\mathbf{w}^g_{r+1}=\sum_{k=1}^{K}\frac{n_k}{n}\mathbf{w}^g_{k,r}
\end{equation}
where the weight $n_k/n$ reflects each client's dataset size. Local parameters $\mathbf{w}^l_k=\{\mathbf{w}^{cls}_k,\mathbf{w}^{adp}_k\}$ are never transmitted, enabling the model to learn a shared quantum-fused multimodal representation while retaining client-specific adaptations for heterogeneous agent environments.

\section{Performance Evaluation}
\label{sec:results}

\begin{table}[b]
\renewcommand{\arraystretch}{1.3}
\centering
\caption{Fusion parameters, per-round training time, and per-client test accuracy (\%) at convergence.}
\label{tab:comparison}
\begin{tabular}{l|cc|cccc|c}
\hline
\textbf{Method} & \textbf{Fusion} & \textbf{Time} & \textbf{S1} & \textbf{S2} & \textbf{S3} & \textbf{S4} & \textbf{Mean} \\
 & \textbf{Params} & \textbf{(s)} & & & & & \\
\hline
Local   & 33K & \textbf{9.8}  & 96.1 & 96.8 & 96.3 & 95.9 & 96.3 \\
FedAvg  & 33K & 10.4          & 96.4 & 97.1 & 96.5 & 96.2 & 96.6 \\
FedProx & 33K & 11.9          & 96.5 & 97.2 & 96.7 & 96.3 & 96.7 \\
MLP-FL  & 33K & 12.5          & 97.2 & 97.8 & 97.1 & 96.8 & 97.2 \\
\hline
QFedAgent & $\mathbf{72}^\dagger$ & 121.4 & \textbf{97.8} & \textbf{98.3} & \textbf{97.5} & \textbf{97.1} & \textbf{97.7} \\
\hline
\multicolumn{8}{l}{$^\dagger$VQC rotation parameters $\boldsymbol{\theta}^{vqc}$; total fusion module is 3,144 parameters} \\
\multicolumn{8}{l}{including $\mathbf{W}_{pre}$ and $\mathbf{W}_{post}$, giving ${\sim}10\times$ reduction over MLP.} \\
\end{tabular}
\end{table}
\subsection{Experimental Setup}

We evaluate QFedAgent on the OPPORTUNITY Activity Recognition dataset, which contains IMU recordings from four subjects (S1--S4) performing daily activities. Each subject is treated as one FL client, forming a natural non-IID partition due to heterogeneous motion patterns. Accelerometer and gyroscope signals from $C=15$ channels are segmented into windows of $T=64$ timesteps with 50\% overlap, producing 2,780--3,760 samples per client across four locomotion classes (Stand, Walk, Sit, Lie). A global test set is constructed from 20\% of all windows, stratified by subject.

The VQC fusion layer uses $N=8$ qubits and $L=3$ \texttt{StronglyEntanglingLayers} with \texttt{AngleEmbedding} (RX rotations), and Pauli-$Z$ expectation values as output. Each CNN encoder produces a $d=128$-dimensional embedding. Federated training runs for $R=10$ rounds with 10 local epochs per round, batch size 16, and the Adam optimizer ($\eta=10^{-3}$) on an NVIDIA RTX 5080 GPU.

We compare against four baselines with the same encoder and classifier: (i) \textit{Local only}; (ii) \textit{FedAvg}~\cite{mcmahan2017}; (iii) \textit{FedProx}~\cite{li2020fedprox} ($\mu=0.01$); and (iv) \textit{MLP-FL}, which replaces the VQC fusion with a classical layer under the same personalized FL protocol.

\subsection{Results}

\begin{figure}[!t]
    \centering
    \includegraphics[width=0.85\linewidth]{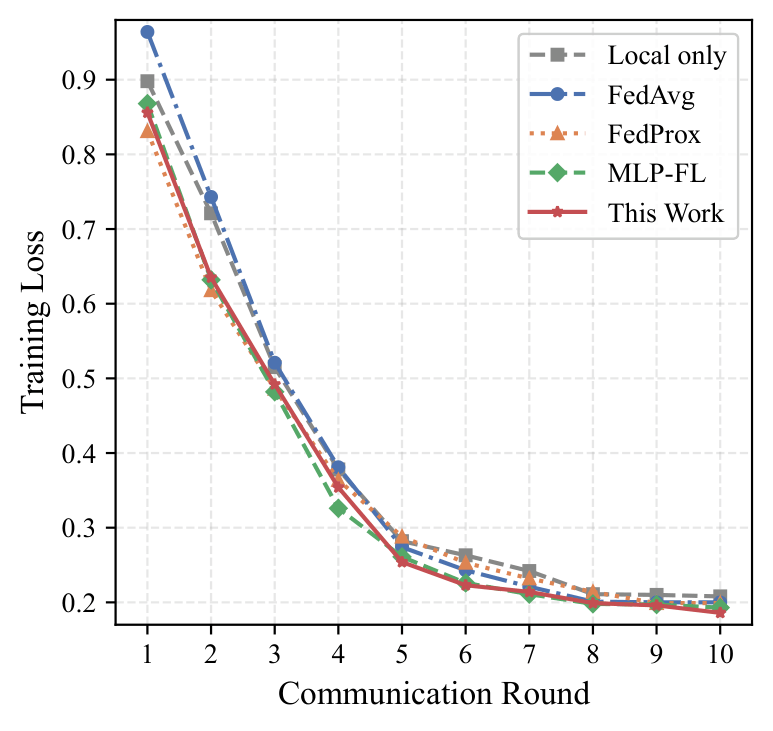}
    \caption{Mean training loss over communication rounds. QFedAgent converges to the lowest final loss (0.186), demonstrating stable federated optimization with quantum-parameterized cross-modal fusion.}
    \label{fig:loss}
\end{figure}

Fig.~\ref{fig:loss} shows the mean training loss over communication rounds. All methods converge rapidly during the first four rounds before stabilizing, confirming that federated optimization remains stable despite non-IID client distributions. Notably, VQC-FL achieves the lowest final loss (0.186), slightly outperforming MLP-FL (0.193). This improvement indicates that the quantum circuit effectively captures cross-modal correlations between accelerometer and gyroscope signals while maintaining stable federated updates. In contrast, FedAvg and the Local-only baseline converge to higher residual losses, highlighting the benefits of personalized training and improved multimodal fusion.

\begin{figure}[!t]
    \centering
    \includegraphics[width=0.95\linewidth]{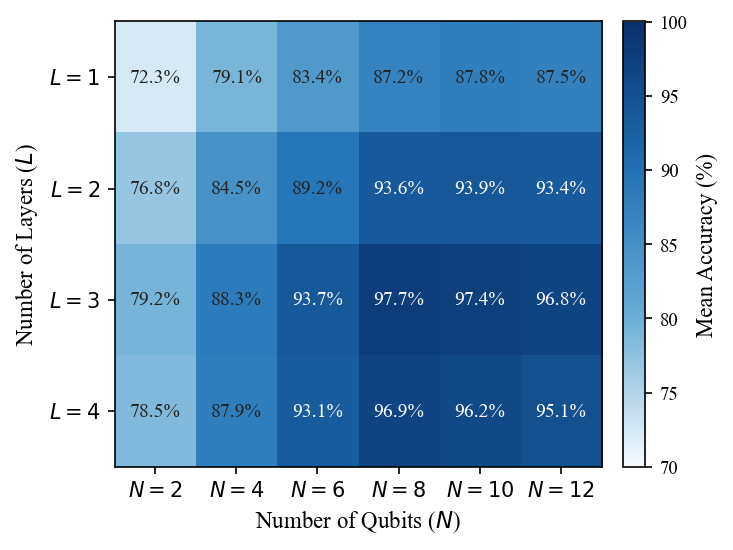}
    \caption{Mean test accuracy (\%) for varying VQC circuit configurations ($N$ qubits, $L$ entangling layers).}
    \label{fig:heatmap}
\end{figure}

Table~\ref{tab:comparison} reports the number of fusion parameters, per-round training time, and per-client test accuracy at convergence. All methods achieve high accuracy (96.3--97.7\%), indicating that the shared CNN encoders provide strong feature representations. Among the classical baselines, MLP-FL reaches the highest mean accuracy of 97.2\%. QFedAgent further improves to 97.7\% while using only 72 quantum rotation parameters versus 33K in MLP-based fusion, achieving approximately $10\times$ total parameter reduction when including classical interface projections. This demonstrates that quantum entanglement can model cross-modal interactions efficiently with far fewer parameters. However, quantum circuit simulation adds significant overhead, increasing per-round training time to 121.4 s versus 9.8--12.5 s for classical methods, as classical hardware must track the full $2^N$-dimensional quantum state and compute parameter-shift gradients with two circuit evaluations per parameter. This cost would not arise on real quantum hardware. Subject S4 consistently exhibits the lowest accuracy due to its smaller dataset (2,780 windows), though it still benefits from federation relative to the Local-only baseline.

Lastly, Fig.~\ref{fig:heatmap} presents an ablation study over circuit configurations. As the number of qubits $N$ and entangling layers $L$ increases, classification accuracy improves steadily up to $N=8$ and $L=3$. Beyond this point, however, improvements diminish while optimization becomes more difficult. This trend suggests that moderate circuit width and depth provide sufficient representational capacity for multimodal fusion while maintaining stable training dynamics.

\section{Conclusion}
\label{sec:conclusion}
This paper presents QFedAgent, a personalized FL framework for multimodal IMU-based activity recognition in multi-agent systems. By replacing classical MLP fusion with a VQC module, QFedAgent achieves approximately $10\times$ total fusion parameter reduction while attaining the highest mean accuracy of 97.7\% over FedAvg, FedProx, and MLP-FL baselines on the OPPORTUNITY dataset. The personalized architecture balances global knowledge transfer with local adaptation under non-IID distributions, demonstrating that parameter-efficient quantum fusion remains effective in federated settings. Although classical simulation incurs higher per-round training time, this overhead is expected to diminish on quantum-ready hardware where circuit execution scales independently of parameter count.

\section*{Acknowledgment}
This research was supported by the Department of Electrical and Computer Engineering, FAMU-FSU College of Engineering, Florida State University.

\bibliographystyle{IEEEtran}
\bibliography{refs}

\end{document}